\documentclass{article}
\usepackage{spconf,amsmath,graphicx}

\usepackage{graphicx}
\usepackage{amsthm}
\usepackage{amssymb,latexsym,multicol,caption}     
\usepackage{algorithm}  
\usepackage{algorithmic}
\usepackage{epsfig} 
\usepackage{epstopdf}
\usepackage[colorlinks,anchorcolor=red,linkcolor=blue,citecolor=red, urlcolor=black]{hyperref}
\usepackage{color}
\usepackage{multirow}
\usepackage{multicol}
\usepackage{subfigure}
\usepackage{placeins}
\usepackage{bm}
\usepackage{graphics}
\usepackage{footnote}
\usepackage{url}
\usepackage{indentfirst}
\usepackage{longtable}
\usepackage{array}
\usepackage{mdwmath}
\usepackage{mdwtab}
\usepackage{rotating}
\usepackage{color}
\usepackage{morefloats}
\usepackage{slashbox}
\usepackage{cite}
\usepackage{mcite}
\usepackage{eqnarray}
\usepackage{float}
\usepackage{makecell}
\usepackage{amsmath}
\usepackage{threeparttable}
\usepackage{pifont}
\usepackage{cleveref}
\usepackage{balance}
\usepackage{flushend}
\usepackage{makecell}

\renewcommand{\eqref}[1]{(\ref{#1})}

\crefname{equation}{}{}
\crefname{figure}{Fig.}{Figs.}
\crefname{table}{Table}{Tables}
\crefname{section}{Section}{Sections}
\crefname{prop}{Proposition}{Propositions}
\crefname{theorem}{Theorem}{Theorems}
\crefname{lemma}{Lemma}{Lemmas}
\crefname{algorithmic}{Algorithm}{Algorithms}

\newtheorem{theorem}{Theorem}[section]

\theoremstyle{plain}

\theoremstyle{definition}

\title{Hyperspectral Image Denoising with Log-based Robust PCA}
\name{Yang Liu$^1$, Qian Zhang$^1$, Yongyong Chen$^2$, Qiang Cheng$^3$, Chong Peng$^{1,*}$}
\address{1. Qingdao University,	2. Harbin Institute of Technology, 3. University of Kentucky}
\begin{document}
%
\maketitle
%
%
\begin{abstract}

	
	It is a challenging task to remove heavy and mixed types of noise from Hyperspectral images (HSIs). 
	In this paper, we propose a novel nonconvex approach to RPCA for HSI denoising, which adopts the log-determinant rank approximation and a novel $\ell_{2,\log}$ norm, 
	to restrict the low-rank or column-wise sparse properties for the component matrices, respectively.
	For the $\ell_{2,\log}$-regularized shrinkage problem, we develop an efficient, closed-form solution, which is named $\ell_{2,\log}$-shrinkage operator, which can be generally used in other problems. 
	Extensive experiments on both simulated and real HSIs demonstrate the effectiveness of the proposed method in denoising HSIs.
	
\end{abstract}





\section{Introduction}
Hyperspectral imaging is widely used in various applications, such as biomedical imaging, terrain classification, and military surveillance \cite{tiwari2011An,zhang2015Compression,zhao2014Hyperspectral}.
However, clean hyperspectral images (HSIs) are rarely obtained due to unavoidable corruptions by mixed types of noise, 
such as Gaussian noise, impulse noise, deadlines, and stripes, in the acquisition process \cite{zhang2014Hyperspectral}, 
which makes it more challenging to process HSIs in various applications, such as classification \cite{li2017Hyperspectral} and unmixing \cite{iordeche2011Sparse}.
Thus, there is an essential demand for developing efficient algorithms to remove noise from HSIs.

In the last decades, a number of HSI denoising techniques have been developed.
For example, \cite{othman2006noise} exploits the dissimilarity of signal regularity in both spatial and spectral dimensions of the HSIs with a hybrid spatial-spectral derivative-domain wavelet shrinkage model;
\cite{zhong2013multiple} propose to simultaneously adopt the spatial and spectral dependences in a unified probabilistic framework;
\cite{qian2013hyperspectral} proposes a sparse representation-based framework that considers both nonlocal similarity and spectral-spatial structure of HSIs. 
Moreover, methods such as principal component analysis, wavelet shrinkage, anisotropic diffusion, and multitask sparse matrix factorization, etc.,
have been considered for HSI denoising \cite{ye2015multitask,chen2011denoising,duarte2007comparative,wang2010anisotropic}.
Most of the above mentioned methods require some specific prior knowledge of the noise.
Unfortunately, the above methods can only remove one or two kinds of noise due to the limitation of prior knowledge on the noise.
Thus, there is still a demand in developing more effective method to remove mixed types of noise from HSIs.

More recently, low-rank matrix recovery-based approach has been developed for HIS denoising \cite{Xu2017Denoising}, which obtains promising performance. 
The basic assumption is that the data can be decomposed into a low-rank and a sparse component,
which generally holds for HSIs with the two components corresponding to clean images and sparse noise, respectively.
Robust principal component analysis (RPCA) has been widely used for low-rank and sparse matrices separation \cite{Cand2011Robust}.
The separation usually relays on the nuclear and $\ell_1$ norm minimization.
Recent studies point out that the nuclear norm may not approximate the true rank function well, 
which may lead to degraded performance in low-rank matrix recovery \cite{peng2015subspace}.
Consequently, some nonconvex approaches to RPCA have been developed to better approximate the rank function, 
which has been shown successful \cite{peng2020robust}. 
The nonconvex RPCA methods mainly focus on developing more accurate low-rank approximation for low-rank matrix recovery.
However, existing methods rarely consider developing more accurate sparse approximation for the sparse matrix recovery.
In fact, there is a close connection between the low-rank and sparse recovery problems. 
The minimization of rank for a matrix actually equals to the minimization of sparseness for all its singular values. 
For nonconvex rank approximations, the improved approximation actually benefits from the improved approximation to the sparsity of the singular values.
Thus, the great success of nonconvex rank approximation inspires us to separate low-rank and sparse matrices with nonconvex approximations to both the rank and sparseness, respectively.

In this paper, we propose a novel RPCA method with log-based approximations to both the rank function and column-wise sparsity. 
We summarize the key contributions of our paper as follows:
1) We propose a novel approximation to approximate column-wise sparsity, named $\ell_{2,\log}$-norm, which is more accurate than the widely used $\ell_{2,1}$ norm. Moreover, the $\ell_{2,\log}$-norm is unitary invariant;
2) We formally provide a closed-form solution to the $\ell_{2,\log}$ norm-based thresholding problem, which can be generally used in various problems that restrict column-wisely sparsity;
3) Efficient optimization algorithm is developed for the proposed model, which is theoretically guaranteed to converge. 
4) We observe superior performance of the proposed method compared with sate-of-the-art baseline methods, which confirm the effectiveness of our method.

\section{Related Work}
\label{sec_related}
Given data matrix $X$, RPCA assumes that the data can be decomposed into a low-rank and a sparse parts, which can be mathematically formed as $X=L+S$.
To do this, the classic RPCA aims at solving the following constrained optimization problem \cite{Cand2011Robust}:
\begin{equation}
\label{eq_rpca_l1}
\min_{L,S} \|L\|_* + \lambda \|S\|_{1},	\quad s.t.\quad X = L+S,
\end{equation}
where $\|\cdot\|_*$ is the nuclear norm that adds all singular values of the input matrix, $\|\cdot\|_1$ is the $\ell_1$ norm that adds the absolute values of all elements of the input matrix, and $\lambda\ge 0$ is a balancing parameter.

\section{RPCA with Log-based Approximations}

Given the observed HSI $\mathcal{X}\in\mathcal{R}^{r\times c\times n}$, with $r$ and $c$ being the spatial dimensions and $n$ being the spectral dimension,
we first divide $\mathcal{X}$ into overlapped patches with each patch has size $q\times q\times n$.
Then we obtain a matrix $X$ of size $q^2\times n$ by lexicographically ordering each patch.
In the rest of this section, we develop our model based on the observation $X$.
It is noted that the new model can be directly applied to the overall data set with reshaped size $(rc)\times n$, 
whereas the patch-based processing follows a common strategy for HSI application.

It is natural that the observed HSI can be decomposed into a low-rank and a sparse component, which correspond to the low-rank clean data and sparse noise, respectively.
To separate them from the data $X$, it is natural to adopt the RPCA model in \cref{eq_rpca_l1}. To enhance the spatial connection information in the sparse component, 
the $\ell_{2,1}$ norm is adopted to replace the $\ell_1$ norm in \cref{eq_rpca_l1}, leading to \cite{xu2012robust}
\begin{equation}
\label{eq_rpca_l21}
\min_{L,S} \|L\|_* + \lambda \|S\|_{2,1}, \quad s.t. \quad X = L+S,
\end{equation}
where $\|S\|_{2,1} = \sum_{j}\|s_j\|_2$ is the $\ell_{2,1}$ norm, and $\|s_j\|_2$ is the $\ell_2$ norm of each column of $S$.

In fact, there is a close connection between the nuclear norm and the $\ell_{2,1}$ norm. 
For a matrix $M$, we define $a$ and $b$ to be vectors that contains singular values of $M$ and $\ell_2$ norm of all columns, respectively.
Then, based on the definitions of the nuclear and $\ell_{2,1}$ norms, it is easily seen that $\|M\|_* = \|a\|_1$ and $\|M\|_{2,1}=\|b\|_1$, respectively.
Thus, the minimization of the nuclear norm and the $\ell_{2,1}$ norm restricts the sparsity of singular values and columns, respectively.
Recently, it is pointed out that the minimization of nuclear norm may not lead to a desired low-rank matrix recovery due to the inaccurate approximation of the nuclear norm to the rank function.
To overcome this issue, nonconvex approaches have been developed to approximate the rank function more accurately than the nuclear norm, 
which has been shown successful in various applications such as subspace clustering \cite{peng2015subspace,peng2016feature} and robust PCA \cite{peng2020robust}. 
This inspires us to adopt nonconvex rank approximation technique to recover the low-rank component from the data.
Specially, we adopt the log-determinant rank approximation, which is defined as 
\begin{equation}
\label{eq_logdet}
h(L) = \operatorname{logdet}(I + (L^TL)^{\frac{1}{2}}) = \sum_{i}(1+\sigma_i(L)),
\end{equation}
where $I$ is an identity matrix of proper size and $\sigma_{i}(\cdot)$ is the $i$-th largest singular value of the input matrix. 
\cref{eq_logdet} leads to the following model:
\begin{equation}
\min_{L,S} \operatorname{logdet}(I+(L^TL)^{\frac{1}{2}}) + \lambda \|S\|_{2,1}, \quad s.t. \quad X = L+S.
\end{equation}
Due to the close connection between the nuclear and the $\ell_{2,1}$ norms, the later suffers from similar issue to the former when there are large values in a matrix.
However, recent works have mainly focused on the nonconvex approximation to the low-rank part while the sparse part is rarely considered.
This inspires us to develop more accurate approximation to restrict the column-wise sparsity for a matrix.
In this paper, to better approximate the column-wise sparsity, we propose the following novel measurement, named $\ell_{2,\log}$ norm:
\begin{equation}
\label{eq_col_log_norm}
\mathcal{C}(S) = \|S\|_{2,\log} = \sum_{i=1}^{n} \operatorname{log}(1+\|s_i\|_2).
\end{equation}
It is seen that $\mathcal{C}(S)$ can measure the column-wise sparsity more accurately than the $\ell_{2,1}$ norm if $S$ contains large values.
Moreover, it is invariant.
Incorporating \cref{eq_col_log_norm} into the objective, we obtain the following Log-based Low-rank and Sparse approximations for RPCA model (LLS-RPCA):
\begin{equation}
\label{eq_obj}
\min_{L,S} h(L) + \lambda \mathcal{C}(S), \quad s.t. \quad X=L+S \\
\end{equation}
For its optimization, we will develop an efficient algorithm based on augmented Lagrange multiplier (ALM) optimization technique.

\section{Optimization}
The augmented Lagrangian function of \cref{eq_obj} is as follows:
\begin{equation}
\begin{aligned}
\mathcal{L} = & h(L) + \lambda \mathcal{C}(S)	+ \frac{\rho}{2}\|X-L-S+\Theta / \rho\|_F^2,
\end{aligned}
\end{equation}
%

\subsection{Optimization w.r.t. $L$}
\label{sec_opt_L}
The sub-problem associated with $L$ is 
\begin{equation}
\label{eq_sub_L}
\min_{L} \operatorname{logdet}(I+(L^TL)^{\frac{1}{2}}) + \frac{\rho}{2}\|X-L-S+\Theta / \rho \|_F^2.
\end{equation}
For a matrix $D$, we define $\mathcal{P}(D)$, $\mathcal{Q}(D)$, and $\sigma_i(D)$ to be its left and right singular vectors and the $i$-th largest singular value, respectively.
Then, similar to \cite{peng2015subspace,peng2020robust}, 
\cref{eq_sub_L} admits a closed-form solution with the following operator:
\begin{equation}
\label{eq_sol_c_uffp}
L = \mathcal{D}_{\frac{1}{\rho}}(X-S+\Theta/\rho),
\end{equation}
where $\mathcal{D}_{ \tau }(D) = \mathcal{P}(D) \text{diag}\{\sigma_i^*\} (\mathcal{Q}(D))^T$, with 
\begin{eqnarray}
\sigma_{i}^* = 
\begin{cases}
\xi,&\mbox{ if $f_i(\xi) \le f_i(0)$ and $ (1 + \sigma_{i}(D))^2 > 4\tau$, }	\\
0, 	&\mbox{ otherwise, }
\end{cases}
\end{eqnarray}
%
where $f_i(x) = \frac{1}{2}(x-\sigma_i(D))^2 + \tau \log (1+x)$, and $\xi = \frac{\sigma_i(D)-1}{2} + \sqrt{\frac{(1+\sigma_i(D))^2}{4} - \tau }$.

\subsection{Optimization w.r.t. $S$}
The sub-problem associated with $S$ is 
\begin{equation}
\label{eq_sub_S}
\begin{aligned}
\min_{S} \lambda \sum_{i=1}^{n} \! \operatorname{log}(1+\|s_i\|_2) \!+\! \frac{\rho}{2}\|X\!-\!L\!-\!S+\Theta / \rho\|_F^2.
\end{aligned}
\end{equation}
For the optimization, we have the following theorem.
\begin{theorem}[$\ell_{2,\log}$-shrinkage operator]
	
	Given matrix $Y\in\mathcal{R}^{d\times n}$ and a nonnegative parameter $\tau$, the following problem
	\begin{equation}
	\label{eq_soft_thres}
	\min_{W} \frac{1}{2} \|Y-W\|_F^2 + \tau \|W\|_{2,\log}
	\end{equation}
	%
	admits closed-form solution in a column-wise manner:
	\begin{equation}
	\label{eq_sol_soft_thres}
	\!w_i \!=\! 
	\begin{cases}
	\frac{\xi}{\|y_i\|_2}\!y_i, \!&\!\!\! \mbox{ \!if $f_i(\xi) \!\le\! \frac{\|y_i\|_2^2}{2}, 
		\frac{(1 \!+\! \|y_i\|_2)^2}{4}\!>\!\tau$, and $\xi >0$}	\\
	0, 	& \!\mbox{ otherwise, }
	\end{cases}
	\end{equation}
	where $f_i(x) = \frac{1}{2}(x-\|y_i\|_2)^2 + \tau \log (1+x),$ and $\xi = \frac{\|y_i\|_2-1}{2} + \sqrt{\frac{(1+\|y_i\|_2)^2}{4} - \tau }.$
	
\end{theorem}
Due to space limit, we do not provide the detailed proof. 
For ease of representation, we denote the $\ell_{2,\log}$-shrinkage operator of \cref{eq_sol_soft_thres} as $\mathcal{F}_{\tau}(Y)$.
Thus, the solution to \cref{eq_sub_S} is
\begin{equation}
S = \mathcal{F}_{\frac{\lambda}{\rho}}(X-L+\Theta/\rho).
\end{equation}

\subsection{Updating $\Theta$ and $\rho$}
For $\Theta$ and $\rho$, we update them in a standard way as follows:
\begin{equation}
\label{eq_update_theta}
\begin{aligned}
\Theta &= \Theta + \rho(X-L-S),	\\
\rho & = \rho\kappa,
\end{aligned}
\end{equation}
where $\kappa>1$ is a parameter that keeps $\rho$ to be increasing at each iteration.  
The optimization theoretically guarantees the convergence of LLS-RPCA. However, due to space limit, we do not expand detailed proof in this paper.

\vspace{-2mm}\section{ Experiments }
In this section, we conduct experiments to testify the performance of the LLS-RPCA in HSI denoising.
In particular, we compare it with RPCA \cite{Cand2011Robust}, LRMR (LRMR) \cite{zhang2014Hyperspectral}, spectral-spatial total variation (SSTV) \cite{dddd2011sstv}, and noise-adjusted LRMA (NAILRMA) \cite{he2015Hyperspectral}.
We test all methods on both synthetic and real-world data sets.
We will present the detailed experimental results in rest of this section.
All experiments are conducted on a Lenovo laptop with Intel Core-i7-9750H CPU @2.6GHz, 16GB RAM, and 64-bit Windows 10 system.

\subsection{Experiments on Simulated Data Sets }
\label{sec_exp_simu}
To quantitatively evaluate the proposed method, we first conduct experiments on synthetic data set. 
We follow \cite{He2015Total} and use the Indian Pine data set \cite{PURR1947} to generate the synthetic data set. 
Totally, we generate 224 bands with each band containing 145$\times$145 pixels. 
Detailed descriptions of the synthetic data set can be found in \cite{He2015Total}. 
We treat the generated data set as the ground truth clean HSIs. 
To evaluate the denoising performance of the LLS-RPCA, we artificially add noise to the generated HSIs. 

\begin{table}
	\caption{ Quantitative Comparison on Synthetic Data Sets }
	\resizebox{0.48\textwidth}{!}
	{
		\begin{tabular}{ |c|c|c|c|c|c|c|}
			\hline
			\multirow{2}{1.5cm}{\centering  }& \multicolumn{6}{c|}{The First Mixed Noise}\\ 
			\cline{2-7}
			\multirow{2}{1.5cm}{}
			& Noisy & SSTV	& LRMR		& NAILRMA				& RPCA				& Ours				\\ \hline
			MPSNR(dB)    	& 16.972&30.431  &33.795	&33.719  &29.711	& $\mathbf{34.823}$		\\	\hline
			MSSIM		& 0.277 &0.774  &0.903	&0.902  & 0.810	&$\mathbf{0.912}$ 		\\	\hline
			ERGAS 		& 332.547 &74.370  &50.087	&50.329  &78.723	&$\mathbf{44.303}$  		\\	\hline
			TIME(s)        &\      &83.686     &205.204   &106.471   &612.642  &$\mathbf{7.671}$\\ \hline
			\multirow{2}{1.5cm}{\centering  }& \multicolumn{6}{c|}{The Second Mixed Noise}\\
			\cline{2-7}
			\multirow{2}{1.5cm}{}
			& Noisy & SSTV	& LRMR		& NAILRMA				& RPCA				& Ours				\\ \hline
			MPSNR(dB)    	& 14.026 &27.998  &30.901	&31.358  &27.377	& $\mathbf{32.126}$ 		\\	\hline
			MSSIM		& 0.199 &0.678  &0.838	&0.840  & 0.731 	&$\mathbf{0.853}$  		\\	\hline
			ERGAS 		& 466.370 &96.043  &68.658	&64.577  &102.107	& $\mathbf{59.176}$		\\	\hline
			TIME(s)        & \      &82.542  &207.204  &105.261 &493.382 &$\mathbf{11.092}$ \\ \hline
		\end{tabular}
	}%
	\label{tab_simulated} 
\end{table}

In particular, we add noise to the clean images in the following two ways:
\begin{itemize}
\item[1)] We fix the noise intensity for all bands. 
	In particular, we add zero-mean Gaussian noise to each band individually, where the variance equals to 0.14. 
	Then, we add fringes to bands 161-190 as follows. 
	For each band, we randomly select 20-40 columns. 
	Then, for each selected column, we add a fixed value to all pixels, where the value is randomly picked from $(-0.25,0.25)$. 
\item[2)] We randomly add noise to each band, where the  SNR value of each band varies from 45 to 55 dB.
	The mean SNR value of all bands is 49.75.
	Then we add salt-and-pepper impulse noise to the data set, where 20\% pixels are randomly corrupted in each band.
	We randomly set the impulse noise intensity of each band from $[0.0196,0.0784]$, where the mean intensity is 0.0492.
\end{itemize}

Then, we apply all methods to the above two noisy data sets.
In our experiments, we adopt three widely used metrics, including including peak SNR (PSNR) \cite{huynh2008Scope},
structural Similarity index (SSIM) \cite{wang2004image}, and ERGAS \cite{wald:hal-00464703}. 
For all methods, we tune their parameters such that best performance can be observed. 
Specially, for RPCA, we use it theoretically optimal value for the balancing parameter.
For each method, we report its averaged performance over all bands in \cref{tab_simulated}. 
It is seen that the proposed method has the best performances in all metrics under both noise settings. 
Among the baseline methods, the LRMR and NAILRMA are the most competing ones. 
Compared with the LRMR (resp. NAILRMA), the proposed method improves the performance in PSNR, SSIM, and ERGAS by 
(1,0.01,6) and (1,0.01,6) (resp. (1.5,0.02,9) and (1,0.01,5)) under the two mixed noise conditions, respectively. 
Compared with other methods, the improvements are more significant. Moreover, the proposed method is significantly faster than SSTV, LRMR and NAILRMA.
These observations confirm the effectiveness of the proposed method from quantitative perspective.

\subsection{Experiments on Real Data Sets }

\begin{figure}[h]
	\centering
	\includegraphics[width=0.9\columnwidth]{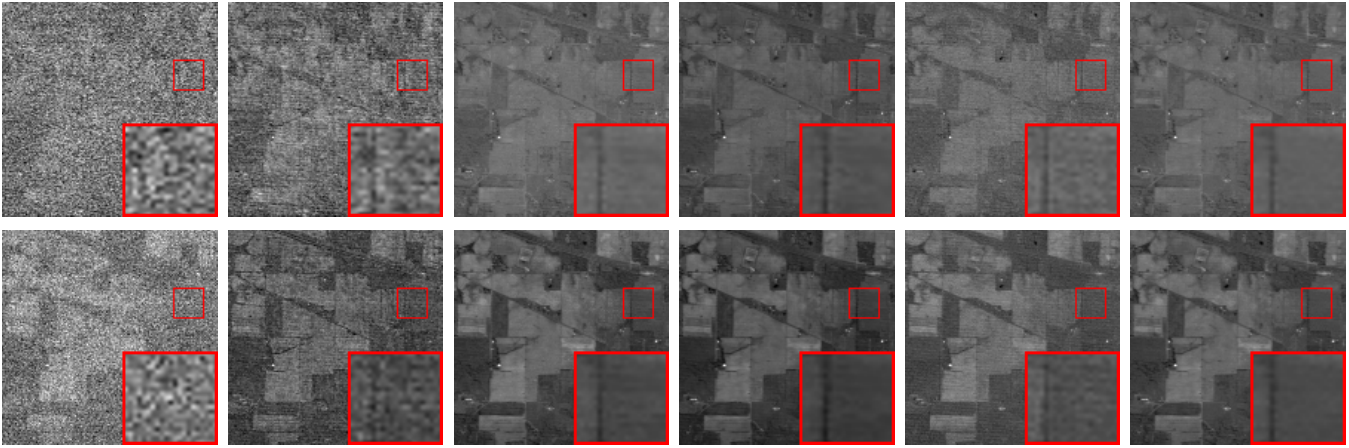}
	\caption{ Restoration results on AVIRIS indian pines data set. From top to bottom are the images located at the 150, and 220th bands, respectively. From left to right are the original and restored images by SSTV, LRMR, NAILRMA, RPCA, and LLS-RPCA, respectively.}
	\label{fig_indian}
\end{figure}

{In this test, we evaluate the proposed method on two real world datasets, including HYDICE unban and AVIRIS Indian Pines. For all methods, we follow the strategy in above test to tune the parameters. Detailed descriptions of the data sets and experimental results are presented in the following.

\subsubsection{HYDICE Urban Dataset}
This data set contains 210 bands of HSIs, where each image has 307$\times$307 pixels.
These images are corrupted by stripes, deadlines, atmosphere, water absorption, and other unknown types of noise.
Some bands in this data set are severely polluted by atmosphere and water absorption, which rarely provide useful information.
Since the real world data set lacks the underlying clean images, we follow a common strategy and visually compare the performance of all methods.
Without loss of generality, we show the results at bands 152, 139, and 207, respectively, 
which involves the bands with both light and heavy noise.
It is observed that the LLS-RPCA can simultaneously remove the mixed types of noise from the noisy images.
It is seen that bands 139 and 207 are heavily corrupted whereas the LLS-RPCA well recovers structural information and preserves rich details. 
Among all baseline methods, NAILRMA has the best performance, which removes almost all noise.
However, some detail information is missing in the recovered images and we can still observe some stripe effects in band 139.
The performance of other baseline methods are less competitive.
For example, all other methods cannot remove stripes in band 139;
These observations suggest effectiveness of the LLS-RPCA and its superior performance to the baseline methods. 

\subsubsection{AVIRIS Indian Pines Dataset}
The AVIRIS Indian Piness dataset is acquired by the Airborne Visible Infrared Imaging Spectrometer (AVIRIS) in Northwestern Indiana in 1992. 
This data set contains images of 145$\times$145 pixels in 220 bands, where some bands are severely damaged by the mixed Gaussian and impulse noise.
We show the performance of these methods in \cref{fig_indian}.
Among all bands, we report the results on the 150 and 220th bands, where it is difficult to observe useful information due to heavy noise.  
It is seen that there are still heavy noises in restored images by the SSTV and RPCA, where local veins are still unrecognizable. 
Among the baseline methods, the LRMR and NAILRMA are the most competitive ones, which remove majority of noise.
However, we can still observe some slight local noises and veins remaining in the smooth regions. 
The LLS-RPCA not only restores the main information image, but also eliminates local noise, which show superior performance to the baseline methods.}



\begin{figure}[h]
	\centering
	\includegraphics[width=0.9\columnwidth]{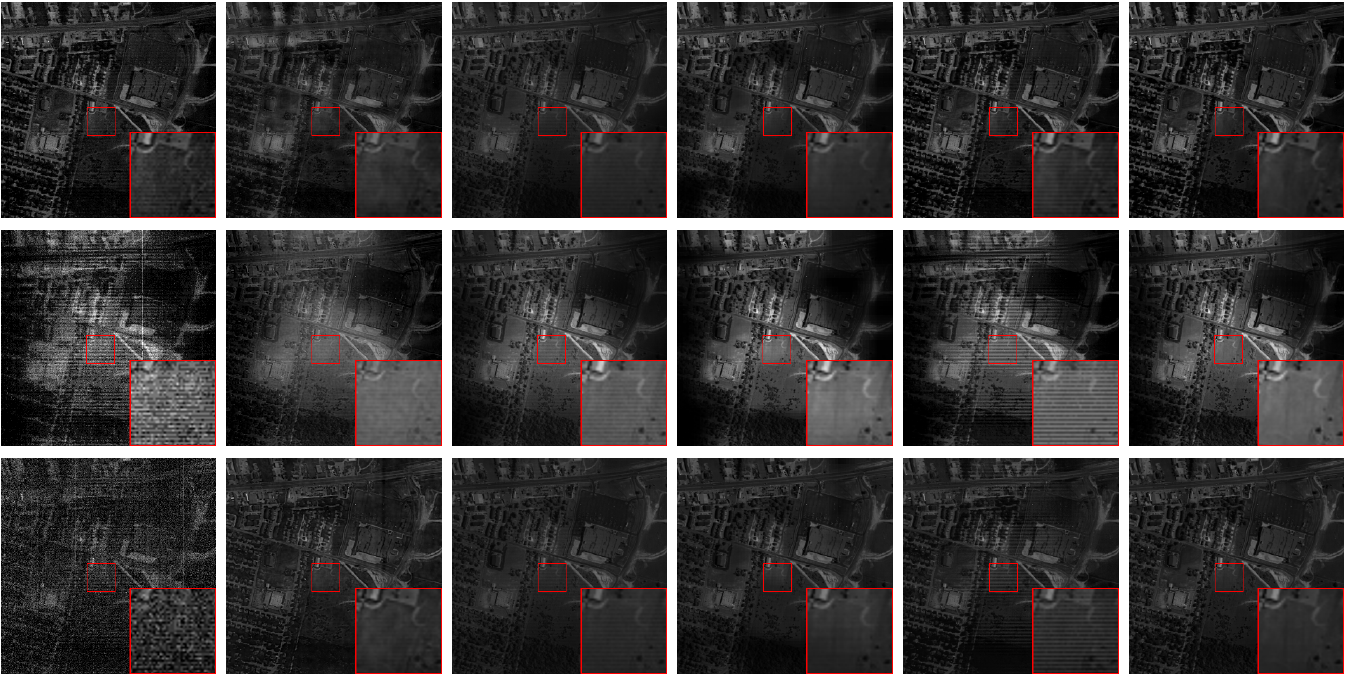} 		\\	 
	\caption{ Restoration results on HYDICE urban data set. From top to bottom are the images located at the 152, 139, and 207th bands, respectively. From left to right are the original and restored images by SSTV, LRMR, NAILRMA, RPCA, and LLS-RPCA, respectively.}
	\label{fig_urban}
\end{figure}

\section{Conclusion}
In this paper, we propose a novel RPCA method, named LLS-RPCA for HSI denoising. 
The new method adopts log-based functions for both low-rank and sparse approximations.
The $\ell_{2,\log}$ norm is more accurate than the widely used $\ell_{2,1}$ norm in approximating column-wise sparsity.
We formally provide solution to $\ell_{2,\log}$-shrinkage problem, which can be generally used for other problems that restrict column-wise sparsity.
Extensive experiments confirm the effectiveness and efficiency of the proposed method in HSI denoising.

{\small\bibliographystyle{IEEEbib}
\bibliography{hsi_logs}}

\end{document}